\documentclass[sigconf]{acmart}
\acmSubmissionID{262}
\usepackage{booktabs} 
\usepackage{subfigure}
\usepackage{xcolor}

\usepackage{changepage}

\usepackage[ruled]{algorithm2e} 

\SetAlFnt{\small}
\SetAlCapFnt{\small}
\SetAlCapNameFnt{\small}
\SetAlCapHSkip{0pt}

\copyrightyear{2022}
\acmYear{2022}
\setcopyright{acmcopyright}\acmConference[SA '22 Conference Papers]{SIGGRAPH Asia 2022 Conference Papers}{December 6--9, 2022}{Daegu, Republic of Korea}
\acmBooktitle{SIGGRAPH Asia 2022 Conference Papers (SA '22 Conference Papers), December 6--9, 2022, Daegu, Republic of Korea}
\acmPrice{15.00}
\acmDOI{10.1145/3550469.3555391}
\acmISBN{978-1-4503-9470-3/22/12}

\usepackage{amsmath,amsfonts,bm}
\usepackage{pifont}

\def\eqref#1{equation~\ref{#1}}

\def\1{\bm{1}}

\def\eps{{\epsilon}}

\def\rva{{\mathbf{a}}}

\def\rvd{{\mathbf{d}}}

\def\rvg{{\mathbf{g}}}

\def\rvu{{\mathbf{i}}}

\def\rvm{{\mathbf{m}}}

\def\rvq{{\mathbf{q}}}

\def\rvs{{\mathbf{s}}}

\def\rvu{{\mathbf{u}}}
\def\rvv{{\mathbf{v}}}

\def\rvx{{\mathbf{x}}}

\def\rvz{{\mathbf{z}}}

\def\rmU{{\mathbf{U}}}

\DeclareMathAlphabet{\mathsfit}{\encodingdefault}{\sfdefault}{m}{sl}
\SetMathAlphabet{\mathsfit}{bold}{\encodingdefault}{\sfdefault}{bx}{n}

\newcommand{\expec}{\mathbb{E}}

\newcommand{\mc}{\mathcal}

\newcommand{\bs}{\backslash}

\newcommand{\ov}{\overline}

\newcommand{\mcI}{\mc{I}}

\newcommand{\mcL}{\mc{L}}

\newcommand{\mcM}{\mc{M}}

\newcommand{\norm}[1]{||#1||}

\newcommand{\tnorm}[1]{\norm{#1}_2}

\newcommand{\set}[1]{\{#1\}}

\newcommand{\jjpar}[1]{\left( #1 \right)}

\newcommand{\jjcases}[1]{\begin{cases} #1 \end{cases}}

\newcommand{\cmark}{\ding{51}}%
\newcommand{\xmark}{\ding{55}}%

\begin{document}
\title{PADL: Language-Directed Physics-Based Character Control}

\author{Jordan Juravsky}
\orcid{0000-0003-2080-7074}
\affiliation{%
 \institution{NVIDIA}
 \institution{University of Waterloo}
 \country{Canada}
 }
\email{jjuravsky@nvidia.com}
\author{Yunrong Guo}
\affiliation{%
 \institution{NVIDIA}
 \country{Canada}
 }
\email{kellyg@nvidia.com}
\author{Sanja Fidler}
\affiliation{%
 \institution{NVIDIA}
 \institution{University of Toronto}
 \country{Canada}
}
\email{sfidler@nvidia.com}
\author{Xue Bin Peng}
\orcid{0000-0002-3677-5655}
\affiliation{%
 \institution{NVIDIA}
  \institution{Simon Fraser University}
 \country{Canada}
}
\email{japeng@nvidia.com}

\begin{abstract}
Developing systems that can synthesize natural and life-like motions for simulated characters has long been a focus for computer animation. But in order for these systems to be useful for downstream applications, they need not only produce high-quality motions, but must also provide an accessible and versatile interface through which users can direct a character's behaviors. Natural language provides a simple-to-use and expressive medium for specifying a user's intent. Recent breakthroughs in natural language processing (NLP) have demonstrated effective use of language-based interfaces for applications such as image generation and program synthesis. In this work, we present PADL, which leverages recent innovations in NLP in order to take steps towards developing language-directed controllers for physics-based character animation. PADL allows users to issue natural language commands for specifying both high-level tasks and low-level skills that a character should perform. We present an adversarial imitation learning approach for training policies to map high-level language commands to low-level controls that enable a character to perform the desired task and skill specified by a user's commands. Furthermore, we propose a multi-task aggregation method that leverages a language-based multiple-choice question-answering approach to determine high-level task objectives from language commands. We show that our framework can be applied to effectively direct a simulated humanoid character to perform a diverse array of complex motor skills.
\end{abstract}



\begin{teaserfigure}
\subfigure[Skill command: "jump and swing sword down".]{\includegraphics[height=0.116\textwidth]{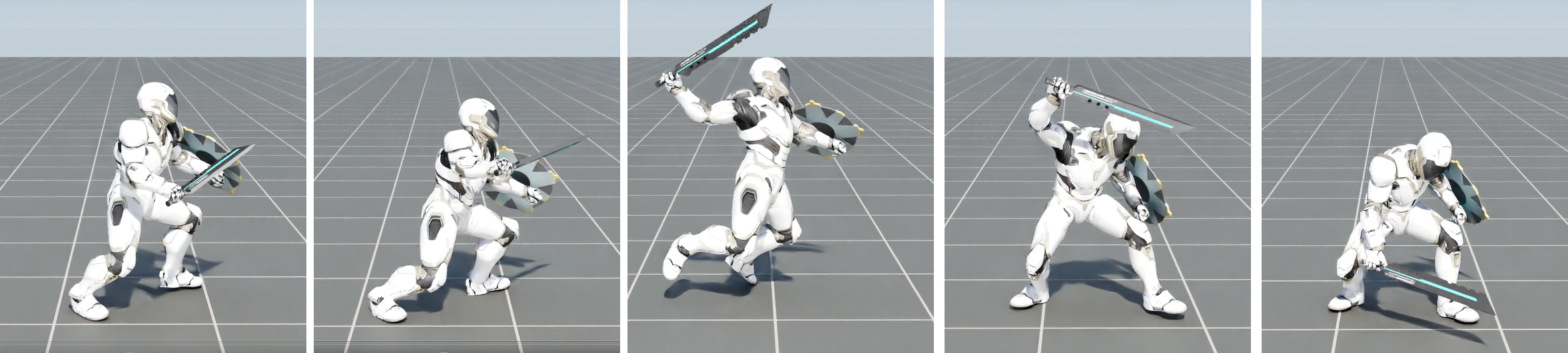}}
\subfigure[Skill command: "shield charge forward".]{\includegraphics[height=0.116\textwidth]{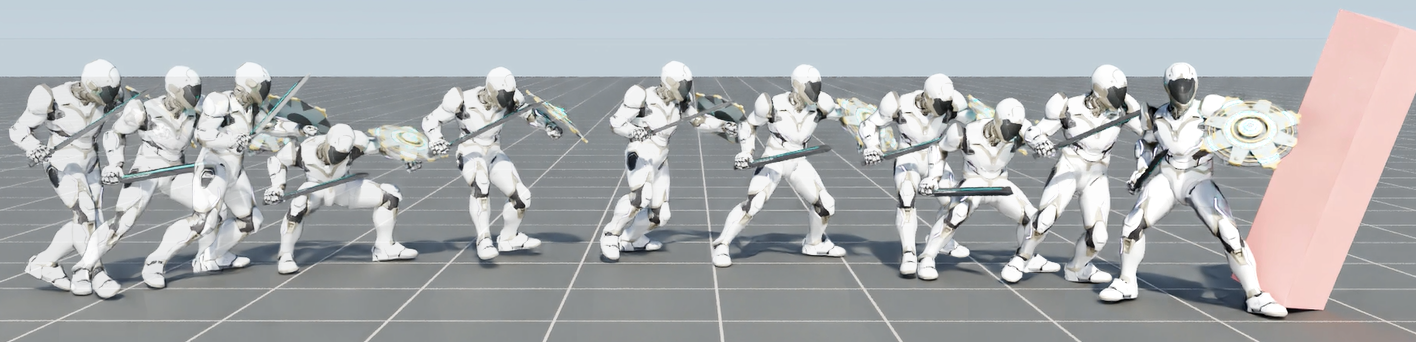}}\\
    \vspace{-0.3cm}
    \caption{Our framework allows users to direct the behaviors of physically simulated characters using natural language commands. \textbf{Left:} Humanoid character performing a jump attack. \textbf{Right:} Character knocking over a target object by performing a shield charge.}
  \label{fig:teaser}
\end{teaserfigure}

\maketitle

\section{Introduction}

Developing physically simulated characters that are capable of producing complex and life-like behaviors has been one of the central challenges in computer animation. Efforts in this domain has led to systems that can produce high-quality motions for a wide range of skills \citep{AthleticsHodgins1995,BipedWang2009,DataDrivenLee2010,delasa2010,CIO2012,BicycleTan2014,2016-TOG-controlGraphs,BasketballLiu2017,DressClegg2018,2018-TOG-deepMimic}. However, in order for these systems to be useful for downstream applications, the control models need not only produce high quality motions, but also provide users with an accessible and versatile interface through which to direct a character's behaviors. This interface is commonly instantiated through compact control abstractions, such as joystick controls or target way points. These control abstractions allow users to easily direct a character's behavior via high-level commands, but they can greatly restrict the variety and granularity of the behaviors that a user can actively control. Alternatively, motion tracking models can provide a versatile interface that enables fine-grain control over a character's movements by directly specifying target motion trajectories. However, authoring motion trajectories can be a labour-intensive process, requiring significant domain expertise or specialized equipment (e.g. motion capture).

An ideal animation system should provide an accessible interface that allows users to easily specify desired behaviors for a character, while also being sufficiently versatile to enable control over a rich corpus of skills. Natural language offers a promising medium that is both accessible and versatile. The recent development of large and expressive language models has provided powerful tools for integrating natural language interfaces for a wide range of downstream applications \cite{bert2018,GPT2020,ClipRadford2021}, such as generating functional code and realistic images from natural language descriptions \cite{Text2Scene2018,Codex2021,Dalle2022}. In this work, we aim to leverage these techniques from NLP to take steps towards developing a language-directed system for physics-based character animation.

The central contribution of this work is a system for language-directed physics-based character animation, which enables users to direct the behaviors of a physically simulated character using natural language commands. Given a dataset of motion clips and captions, which describe the behaviors depicted in each clip, our system trains control policies to map from high-level language commands to low-level motor commands that enable a character to reproduce the corresponding skills. We present an adversarial imitation learning approach that allows a policy to reproduce a diverse array of skills, while also learning to ground each skill in language commands. Our policies can also be trained to perform additional auxiliary tasks. We present a language-based multi-task aggregation model, which selects between a collection of task-specific policies according to a given command, thereby allowing users to easily direct a character to perform various high-level tasks via natural language. 
We present one of the first systems that can effectively leverage language commands to direct full-body physically simulated character to perform a diverse array of complex motor skills. The code for this work is available at \href{https://github.com/nv-tlabs/PADL}{https://github.com/nv-tlabs/PADL}.

\section{Related Work}

Synthesizing natural and intelligent behaviors for simulated characters has been a core subject of interest in computer animation, with a large body of work focused on building kinematic and physics-based control models that can generate life-like motions \citep{AthleticsHodgins1995,Silva2008SimulationOH,BipedWang2009,DataDrivenLee2010,BicycleTan2014,MuscleWang2012,AnimHolden2016,BasketballLiu2017,DressClegg2018}. While a great deal of emphasis has been placed on motion quality, considerably less attention has been devoted on the \emph{directability} of the resulting models at run-time. Directability is often incorporated into these models via control abstractions that allow users to direct a character's behaviors through high-level commands. These abstractions tend to introduce a trade-off between accessibility and versatility. Simple control abstractions, such as joystick commands or target waypoints, \citep{Treuille2007,Coros09,MotionFields2010,taskBasedLocomotion2016,PFNN2017,2018-TOG-deepMimic,NSM2019,AmorphousZhang2020,2020-TOG-MVAE,2021-TOG-AMP,TimeCritical2021,Lee2021Parameterized,2022-TOG-ASE}, provide an accessible interface that can be easily adopted by users. But these abstractions can also limit the versatility of the behaviors that can be actively controlled by a user. Alternatively, general motion tracking models can provide a versatile interface, which allows for fine-grain control over a character's movements through target motion trajectories \citep{Pollard2002,Yamane2010,PredictSimPark2019,DreCon2019,wang2020unicon,ScalableWon2020}. These target trajectories specify desired poses for the character to reach at every timesteps, which in principle can direct the character to perform any feasible motion. However, this versatility often comes at the cost of accessibility, since authoring target motion trajectories can be as tedious and labour intensive as manual keyframe animation. Motion capture can be a more expeditious approach for generating target trajectories for motion-tracking models \citep{2018-TOG-SFV,wang2020unicon,SimPoE2021,DynamicYu2021}, but tends to require specialized equipment and may limit the reproducible behaviors to those that can be physically performed by the user.
In this work, we aim to leverage natural language to develop an accessible and versatile control interface for physics-based character animation.

\paragraph{Natural Language Processing:}
Language models trained on increasingly large datasets have been shown to develop powerful representations for text data~\citep{bert2018, robertaLiu, t5Raffel}, which can be used for a wide range of downstream applications.
One such example is text-guided synthesis, where a user's prompt, expressed in natural language, can be used to direct models to produce different types of content. Large autoregressive models are able to generate coherent text completions given a user's starter prompt \citep{GPT2020}. These models lead to the popularization of ``prompt engineering", where the aim is to construct optimal prompt templates that elicit the desired behaviors from a language model. Such prompt-based systems, often combined with filtering or other post-processing techniques, have been successfully used to solve grade-school math problems and competitive programming challenges \citep{openAIGradeSchoolMath, alphaCode}. Text-guided synthesis can also be applied across different modalities. Here, the language model does not directly generate the desired content, instead it provides a semantically meaningful encoding for a user's language prompt, which can then be used by a separately trained decoder to generate content in a different modality. \citet{glideNichol} and \citet{Dalle2022} successfully used this approach to generate photo-realistic images from natural language, leveraging the text encoder from CLIP \cite{ClipRadford2021}.
In this work, we aim to leverage  powerful language models to develop language-directed controllers for physics-based character animation.

\paragraph{Language-Directed Animation:}
Synthesizing motion from language is one of the core challenges of audio-driven facial animation, where the goal is to generate facial motions for a given utterance. These models typically take advantage of the temporal correspondence between units of speech (phonemes) and facial articulations (visemes) in order to synthesize plausible facial animations for a particular utterance \citep{Pelachaud1996,VoicePuppetry1999,Hong2002,SpeechAnimDeena2009,Audio2Face2017}. A similar temporal correspondence can also be leveraged to generate full-body gestures from speech \citep{BodyLanguageLevine2009,SpeechGesture2020,Language2Pose2019}. While these techniques can be highly effective for generating realistic motions from speech, they are not directly applicable in more general settings where there is no clear temporal correspondence between language and motion. For example, a high-level command such as ``knock over the red block'' implicitly encodes a sequence of skills that a character should perform. 
Sequence-to-sequence models have been proposed to map high-level language descriptions to motion trajectories \citep{PlappertMA17,linvigil18}. \citet{Language2Pose2019} and \citet{MotionClipTevet2022} proposed autoencoder frameworks that learns a joint embedding of language and motion, which can be used to generate full-body motions from language descriptions. While these techniques have demonstrated promising results, they have been primarily focused on developing kinematic motion models. In this work, we aim to develop a language-directed model for physics-based character animation, which maps high-level language commands to low-level controls that enable a character to perform the desired behaviors.

\section{Background}
Our characters are trained using a goal-conditioned reinforcement learning framework, where an agent interacts with an environment according to a control policy $\pi$ in order to fulfill a given goal $\rvg \in \mathcal{G}$, drawn from a goal distribution $\rvg \sim p(\rvg)$. At each time step $t$, the agent observes the state of the environment $\rvs_t \in \mathcal{S}$, and responds by applying an action $\rva_t \in \mathcal{A}$, sampled from the policy $\rva_t \sim \pi(\rva_t | \rvs_t, \rvg)$. After applying the action $\rva_t$, the environment transitions to a new state $\rvs_{t+1}$, and the agent receives a scalar reward $r_t = r(\rvs_t, \rva_t, \rvs_{t+1}, \rvg)$ that reflects the desirability of the state transition for the given goal $\rvg$. The agent's objective is to learn policy $\pi$ that maximizes its expected discounted return $J(\pi)$,
\begin{align}
    J(\pi) = \expec_{p(\rvg)} \expec_{p(\tau | \pi, \rvg)} \left[ \sum_{t=0}^{T - 1} \gamma^t r_t\right],
\end{align}
where $p(\tau | \pi, \rvg) = p(\rvs_0) \prod_{t=0}^{T-1} p(\rvs_{t+1} | \rvs_t, \rva_t) \pi(\rva_t | \rvs_t, \rvg)$ denotes the likelihood of a trajectory $\tau = (\rvs_0, \rva_0, \rvs_1, ..., \rvs_T)$ under a policy $\pi$ given a goal $\rvg$, $p(\rvs_0)$ is the initial state distribution, and $p(\rvs_{t+1} | \rvs_t, \rvs_a)$ represents the transition dynamics of the environment. $T$ is the time horizon of a trajectory, and $\gamma \in [0, 1]$ is a discount factor.

\section{Overview}

In this paper we introduce Physics-based Animation Directed with Language (PADL; pronounced ``paddle''), a system for developing language-directed control models for physics-based character animation. Our framework allows users to control the motion of a character by specifying a \emph{task} to complete, as well as a specific \emph{skill} to use while completing that task. Tasks represent high-level objectives that the agent must accomplish, such as navigating to a target location or interacting with a specific object. In addition to specifying \textit{what} task an agent must accomplish, it is important for users to be able to control \textit{how} the task is accomplished. For example, given the task of navigating to a target location, an agent can walk, run, or jump to the target. In our system, the desired task and skill for the character are specified separately via natural language in the form of a task command and a skill command.

\begin{figure}
    \centering
    \includegraphics[width=0.9\columnwidth]{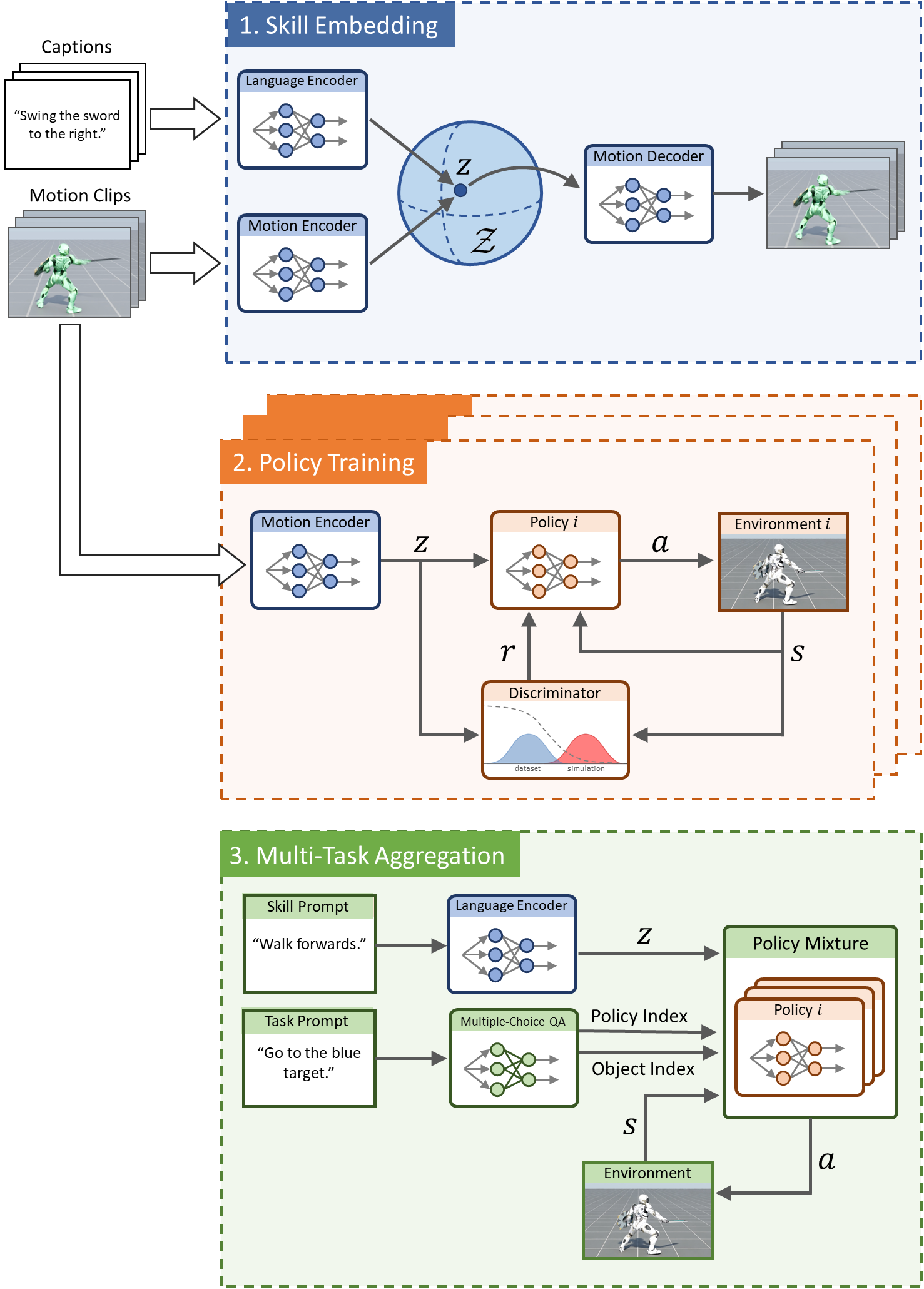}
    \caption{The PADL framework consists of three stages. 1) In the Skill Embedding stage, a dataset of motion clips and corresponding text captions are used to learn a joint embedding of motions and captions. 2) In the Policy Training stage, the learned skill embedding is used to train a collection of policies to perform various tasks, while imitating behaviors in the dataset. 3) Finally, in the Multi-Task Aggregation stage, policies trained for different tasks are combined into a multi-task controller that can be directed to perform different tasks and skills via language commands.}
    \vspace{-0.4cm}
    \label{fig:overview}
\end{figure}

Our framework consists of three stages, and a schematic overview of the system is available in Figure \ref{fig:overview}. First, in the \emph{Skill Embedding} stage, a reference motion dataset $\mcM = \set{(\rvm^i, c^i)}$, containing motion clips $\rvm^i$ annotated with natural language captions $c^i$, is used to learn a shared embedding space $\mathcal{Z}$ of motions and text. Each motion clip $m^i = \{\hat{\rvq}^i_t \}$ is represented by a sequence of poses $\hat{\rvq}^i_t$. A motion encoder $z_m^i = \mathrm{Enc}_m(\rvm^i)$ and language encoder $z_l^i = \mathrm{Enc}_l(c^i)$ are trained to map each motion and caption pair to similar embeddings $z_m^i \approx z_l^i$. Next, in the \emph{Policy Training} stage, this embedding is used to train a collection of reinforcement learning policies, where each policy $\pi^i(\rva_t| \rvs_t, \rvg, \rvz)$ is trained to perform a particular task using various skills $\rvz \in \mathcal{Z}$ from the embedding. Once trained, the policy can then be directed to execute a particular skill by conditioning $\pi$ on the embedding of a given language command $z_l = \mathrm{Enc}_l(c)$. Finally, in the \emph{Multi-Task Aggregation} stage, the different policies are integrated into a multi-task controller that can be directed using language commands to perform a specific task using a desired skill.

\section{Skill Embedding}
\label{sec:skill-embedding}

In the Skill Embedding stage, our objective is to construct an embedding space that aligns motions with their corresponding natural language descriptions. To do this, we follow a similar procedure as MotionCLIP \citep{MotionClipTevet2022}, where a transformer autoencoder is trained to encode motion sequences into a latent representation that ``aligns'' with the language embedding from a pre-trained CLIP text encoder \citep{ClipRadford2021}. Given a motion clip $\hat{\rvm} = (\hat{\rvq}_1, ..., \hat{\rvq}_n)$ and its caption $c$, a motion encoder $\rvz = \mathrm{Enc}_m(\hat{\rvm})$ maps the motion to an embedding $\rvz$. The embedding is normalized to lie on a unit sphere $||\rvz|| = 1$. Following~\citet{MotionClipTevet2022}, $\mathrm{Enc}_m\left(\rvm \right)$ is modeled by a bidirectional transformer \citep{bert2018}. A motion decoder is jointly trained with the encoder to produce a reconstruction sequence $\rvm = (\rvq_1, ..., \rvq_n)$ to recover $\hat{\rvm}$ from $\rvz$. The decoder is also modelled as a birectional transformer $\rvm = \mathrm{Dec}(\rvz, \rmU)$, which decodes all frames of in parallel using a learned constant query sequence $\rmU = (\rvu_1, ..., \rvu_n)$, similar to the final layer of \citet{detr}. The autoencoder is trained with the loss:

\begin{align}
\mcL_{\text{auto}} = \mcL_{\text{recon}} + 0.1\mcL_{\text{align}} .
\end{align}
The reconstruction loss $\mcL_{\text{recon}}$ measures the error between the reconstructed sequence and original motion:
\begin{align}
\mcL_{\text{recon}} = \frac{1}{n} \sum_{t=1}^n \tnorm{\hat{\rvq}_t - \mathrm{Dec}\left(\mathrm{Enc}_m\left(\hat{\rvm} \right), \rmU \right)}^2 .
\end{align}
The alignment loss $\mcL_{\text{align}}$ measures the cosine distance between a motion embedding and the language embedding:
\begin{align}
\mcL_{\text{align}} = 1 - d_{\text{cos}}\left(\mathrm{Enc}_m\left(\hat{\rvm}\right), \mathrm{Enc}_l(c)\right) .
\label{eqn:loss_align}
\end{align}
The language encoder $\mathrm{Enc}_l\left(\rvm \right)$ is modeled using a pre-trained CLIP text encoder with an added head of two fully-connected layers, where only this output head is fine-tuned according to Eq.~\ref{eqn:loss_align}. To help avoid overfitting, for every minibatch of motion sequences sampled from the dataset we also extract a random subsequence from each motion and add these slices to the batch that the model is trained on. These subsequences only contribute to the reconstruction loss.

\section{Policy Training}
\label{sec:policy-training}

Once we have a joint embedding of motions and captions, we will next use the embedding to train control policies that enable a physically simulated character to perform various high-level tasks while using skills specified by language commands. At each timestep $t$, the policy $\pi(\rva_t | \rvs_t, \rvg, \rvz)$ receives as input the state of the character $\rvs_t$, a task-specific goal $\rvg$, and a skill latent $\rvz$. The goal $\rvg$ specifies high-level task objectives that the character should achieve, such as moving to a target location or facing a desired direction. The skill latent $\rvz$ specifies the skill that the character should use to achieve the desired goal, such as walking vs running to a target location. The latents are generated by encoding motion clips $\rvz = \mathrm{Enc}_m(\rvm)$ sampled from the dataset $\mcM$. In order to train a policy to perform a given task using a desired skill, we utilize a reward function consisting of two components:

\begin{equation}
    r_t = r^\text{skill}_t + \lambda^\text{task} r^\text{task}_t,
\end{equation}
where $r^\text{skill}_t$ is a skill-reward, and $r^\text{task}_t$ is a task-reward with coefficient $\lambda^\text{task}$.

\subsection{Skill Objective}

To train the policy to perform the skill specified by a particular $\rvz_i$, we enforce that the policy's distribution of state transitions $(\rvs, \rvs')$ matches that of the corresponding motion clip $\rvm^i$. To accomplish this, we train an adversarial discriminator $D(\rvs, \rvs', \rvz)$ on the joint distribution of state transitions and skill encodings \citep{GAIL2016,MerelTTSLWWH17,2021-TOG-AMP}. The discriminator is trained to predict if a given state transition $(\rvs, \rvs')$ is from the motion clip corresponding to $\rvz$, or if the transition is from the simulated character or from a different motion clip in the dataset. The discriminator is trained by minimizing the following loss:
\begin{align}
\mcL_{D} = \expec_{p_\mathcal{M}(\rvm)} &  \bigg[ -\expec_{p_\rvm(\rvs, \rvs')} \left[\log(D(\rvs, \rvs', \rvz)) \right] \\
&- w_{D} \ \expec_{p_\pi(\rvs, \rvs' |\rvz)} \left[\log(1 - D(\rvs, \rvs', \rvz)) \right]\\
&- (1 - w_{D}) \ \expec_{p_{\mcM \bs \rvm}(\rvs, \rvs')} \left[ \log(1 - D(\rvs, \rvs', \rvz)) \right] \\
&+ w_\mathrm{gp} \ \expec_{p_\rvm(\rvs, \rvs')} \left[ \left| \left| \nabla_\phi D(\phi, \rvz) \middle|_{\phi = (\rvs, \rvs')}\right| \right|^2 \right]\bigg].
\end{align}
$p_\mathcal{M}(\rvm)$ represents the likelihood of sampling a motion clip $\rvm$ from a dataset $\mathcal{M}$, and $\rvz = \mathrm{Enc}_m(\rvm)$ is the encoding of the motion clip. $p_\rvm(\rvs, \rvs')$ denotes the likelihood of observing a state transition from a given motion clip, and $p_\pi(\rvs, \rvs' | \rvz)$ is the likelihood of observing a state transition from the policy $\pi$ when conditioned on $\rvz$. $p_{\mcM \bs \rvm}(\rvs, \rvs')$ represents the likelihood of observing a state transition by sampling random transitions from other motion clips in the dataset, excluding $\rvm$, and $w_D$ is a manually specified coefficient. The final term in the loss is a gradient penalty with coefficient $w_\mathrm{gp}$ \citep{2021-TOG-AMP}, which improves stability of the adversarial training process. The skill-reward is then given by:
\begin{align}
r^\mathrm{skill}_t = -\mathrm{log}\left(1 - D(\rvs_t, \rvs_{t+1}, \rvz) \right).
\end{align}

To direct the policy with a skill command $c_\text{skill}$ after it has been trained, the model is provided with the encoding $\rvz = \mathrm{Enc}_l(c_{\text{skill}})$. By conditioning the discriminator on both state transitions and latents, our method explicitly encourages the policy to imitate every motion clip in the dataset, which can greatly reduce mode collapse. We elaborate on this benefit and compare our approach to related adversarial RL frameworks in Appendix~\ref{sec:mode-collapse}.

\section{Multi-Task Aggregation}

Each policy from the Policy Training stage is capable of performing a variety of skills, but each is only able to perform a single high-level task involving a single target object. We show that these individual policies can be aggregated into a more flexible composite policy, which allows users to direct the character to perform a variety of different tasks in an environment containing multiple objects. However, in our experiments, we found that attempting to use the procedure in Section \ref{sec:policy-training} to train a single multi-task policy to perform all tasks leads to poor performance. Effectively training multi-task policies remains a challenging and open problem in RL, and prior systems have often taken a divide-and-conquer approach for multi-task RL \citep{MultiTaskRuder2017,ghosh2018divideandconquer,PolicyDistillation2015}.

To create a more flexible multi-task, multi-object controller, we aggregate a collection of single-task policies together. At each timestep, the user's current task command is used to generate prompts that are fed to a multiple-choice question-answering (QA) model. The QA model identifies which task  and environment object are being referenced by the user. The single-task controller for the identified task is then set as the active policy controlling the character, and the state of the identified object is passed to the selected policy. An overview of this procedure is provided with pseudocode in Algorithm \ref{alg:multi-task} in the Appendix. Note that since the character is being driven by a single policy from Section \ref{sec:policy-training} at every timestep, the aggregated controller can only follow one high-level task involving a single object at a time. However, with this controller the user can dynamically control which task and object are focused on using natural language.

\subsection{Multiple Choice Question Answering}

\begin{figure}
    \centering
    \includegraphics[width=\columnwidth]{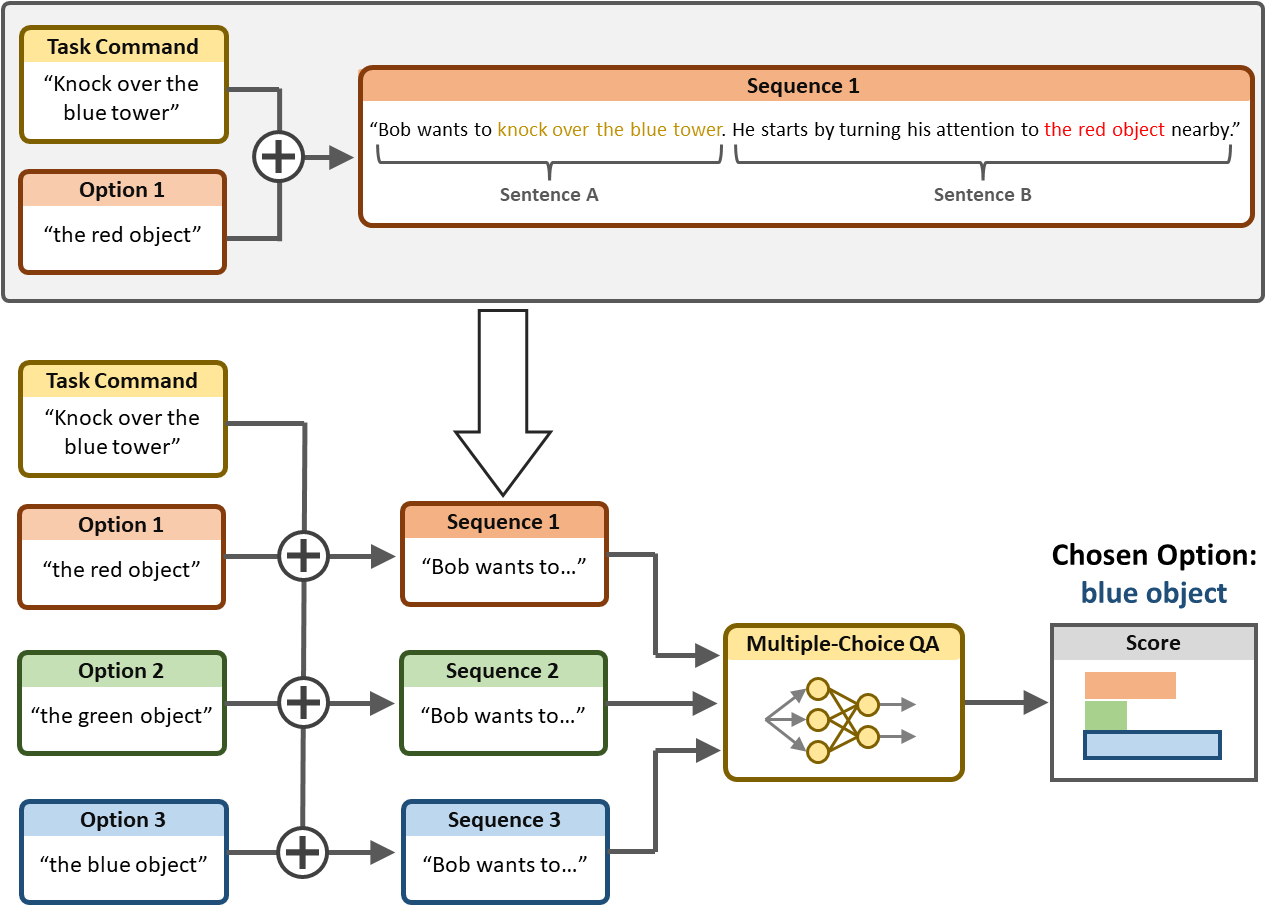}
    \caption{Overview of the language-based selection model used to select a target object based on the user's task command. The task command is used to generate a collection of candidate sentences, each corresponding to a particular object in the environment. A multiple-choice QA model is then used to predict the most likely candidate sentence, based on the task command. The model's prediction is used to identify the target object the user referenced.}
    \vspace{-0.5cm}
    \label{fig:taskCommand}
\end{figure}

An overview of the language-based selection model is shown in Figure~\ref{fig:taskCommand}. The multiple-choice QA model is constructed using a pre-trained BERT model fine-tuned on the SWAG dataset \citep{swagDataset}. Each multiple-choice question is formulated as an initial prompt sentence (Sentence A) alongside $n$ candidate follow-up sentences (Sentence B) \citep{bert2018}. The model then outputs scores for $n$ distinct sequences, where sequence $i$ is the concatenation of the prompt sentence with the $i$-th candidate sentence. The object corresponding to the candidate sentence with the highest score is selected as the target object for the policy. A similar process is used to identify the task from the user's command. 

For each task command provided by the user, the model is provided with two separate multiple-choice questions to identify the relevant task and object, respectively. The first question identifies the task, where each multiple choice option corresponds to a trained policy. The inputs to the QA model follow a story-like format in order to mimic the elements of the SWAG dataset that the model was fine-tuned on. For example, if the task command is \emph{``knock over the blue tower''}, the candidate sequence for the strike policy is:
\begin{itemize}
    \item "Bob wants to \underline{\emph{knock over the blue tower}}. This should be easy for him since he possesses the ability to \underline{\emph{knock over a specified}} \underline{\emph{object}}."
\end{itemize}
Similarly, the candidate sequence for the location policy is given by:
\begin{itemize}
    \item "Bob wants to \underline{\emph{knock over the blue tower}}. This should be easy for him since he possesses the ability to \underline{\emph{navigate to a specified}} \underline{\emph{destination}}."
\end{itemize}
The multiple-choice QA model will then predict which sequence of sentences are most likely. Similarly, in the multiple-choice question to extract the target object, each object is given a multiple choice option describing the object's appearance. The candidate sequence for the green block is given by:
\begin{itemize}
    \item "Bob wants to \underline{\emph{knock over the blue tower}}. He starts by turning his attention to \underline{\emph{the green object}} nearby."
\end{itemize}

\section{Experimental Setup}

\begin{figure*}[t]
	\centering
    \subfigure["sprint forwards while swinging arms".]{\includegraphics[width=\linewidth]{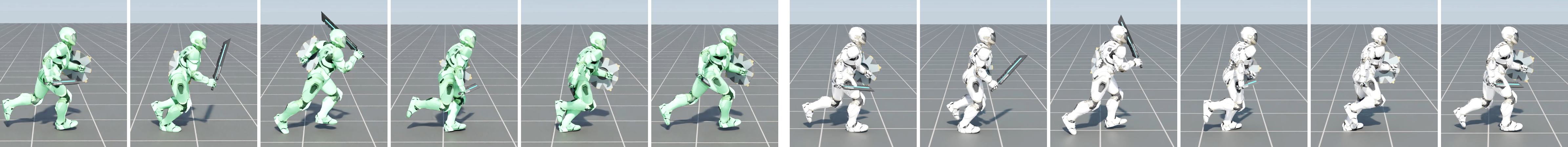}}\\
    \vspace{-0.2cm}
    \subfigure["left shield bash", "shield bash left", "shield bash to the left while standing still".]{\includegraphics[width=\linewidth]{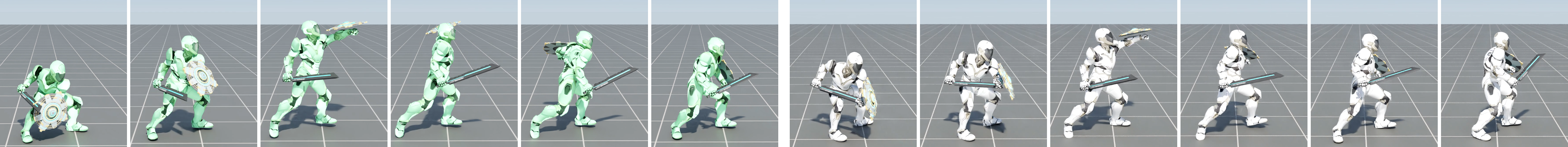}}\\
    \vspace{-0.2cm}
    \subfigure["slash right", "right swing", "swing sword to the right", "stand still and slash to the right".]{\includegraphics[width=1\linewidth]{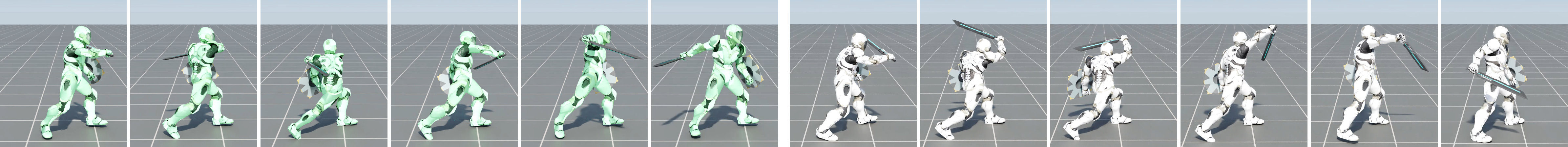}}\\
    \vspace{-0.2cm}
	\subfigure[task: Location. skill: "sprint forward while swinging arms".]{\includegraphics[height=0.114\linewidth]{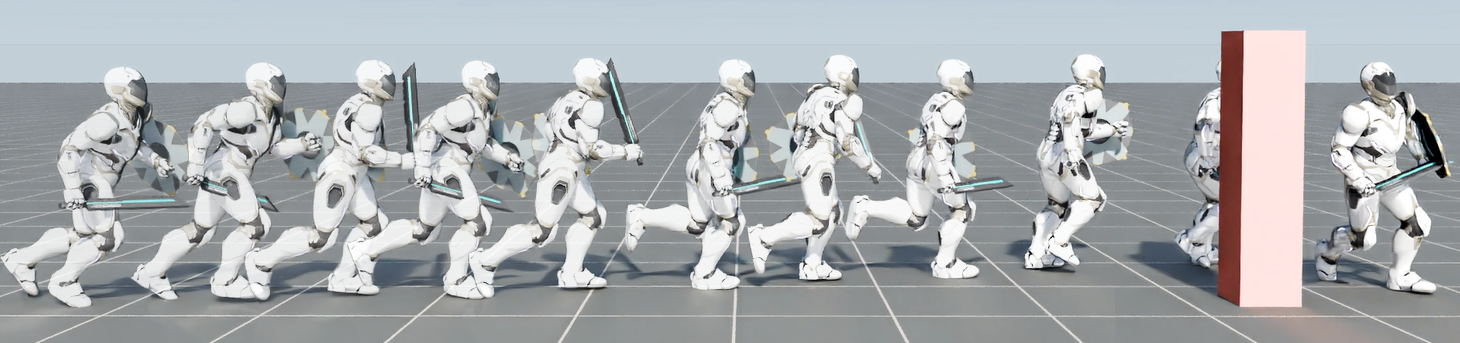}}
   \subfigure[task: Strike. skill: "shield bash to the right".]{\includegraphics[height=0.114\linewidth]{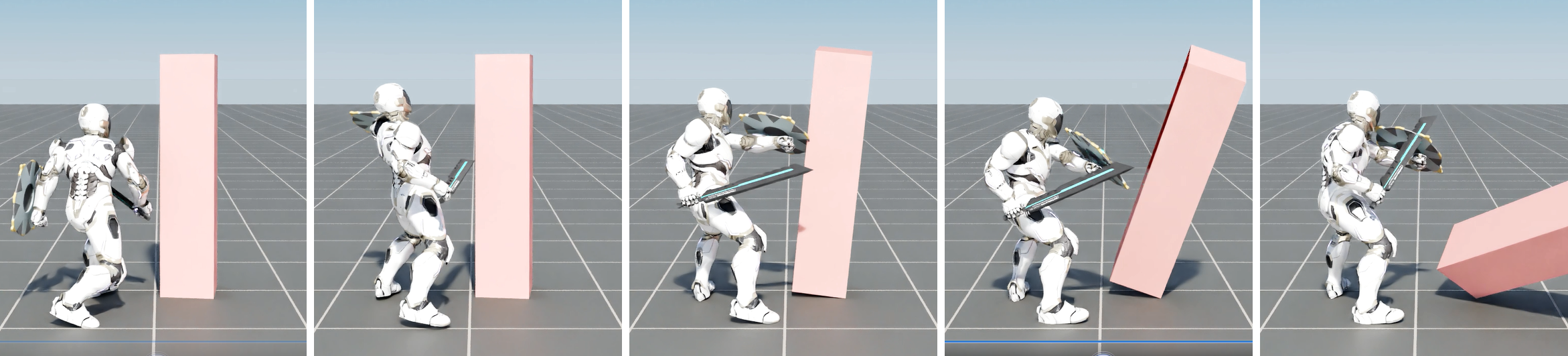}}\\
    \vspace{-0.4cm}
    \caption{(a) -- (c): Reference motion clips (left side) and their corresponding captions, along with motions produced by a simulated character when directed to perform the reference skills through language commands (right side). More reference motions and policy trajectories are shown in Fig.~\ref{fig:moreFilmstripSkills} in the Appendix. (d) -- (e): Trained policies completing tasks with different skills.}
    \label{fig:filmstripSkills}
    \vspace{-0.4cm}
\end{figure*}

We evaluate the effectiveness of our framework by training language-directed control policies for a 3D simulated humanoid character. The character is equipped with a sword and shield, similar to the one used by \citet{2022-TOG-ASE}, with 37 degrees-of-freedom, and similar state and action representations. The dataset contains a total of 131 individual clips, for a total of approximately 9 minutes of motion data. Each clip is manually labeled with 1-4 captions that describe the behavior of the character within a particular clip, for a total of 265 captions in the entire dataset. Fig.~\ref{fig:filmstripSkills} illustrates examples of motion clips in the dataset along with their respective captions. 

\subsection{Tasks}
\label{sec:tasks}

In addition to training policies to imitate skills from the dataset, each policy is also trained to perform an additional high-level task.  Here, we provide an overview of the various tasks, and more detailed descriptions are available in Appendix~\ref{app:tasks}.

\begin{enumerate}
    \item \textbf{Facing:} First, we have a simple facing task, where the objective is for the character to turn and face a target direction $\rvd^*$, encoded as a 2D vector on the horizontal plane. The goal input $\rvg_t = \tilde{\rvd}^*_t$ for the policy records the goal direction in the character's local coordinate frame.
    \item \textbf{Location:} Next, we have a target location task, where the objective is for the character to navigate to a target location $\rvx^*$. The goal $\rvg_t = \tilde{\rvx}^*_t$ records the target location in the character's local coordinate frame $\tilde{\rvx}^*_t$.
    \item \textbf{Strike:} Finally, we have a strike task, where the objective is for the character to knock over a target object. The goal $\rvg_t = (\tilde{\rvx}^*_t, \tilde{\dot{\rvx}}^*_t, \tilde{q}^*_t, \tilde{\dot{q}}^*_t)$ records the target object's position $\tilde{\rvx}^*_t$, rotation $\tilde{q}^*_t$, linear velocity $\tilde{\dot{\rvx}}^*_t$, and angular velocity $\tilde{\dot{q}}^*_t$. All features are expressed in the character's local frame.
\end{enumerate}

\subsection{Training}

All physics simulations are performed using Isaac Gym, a massively parallel GPU-based physics simulator \citep{IsaacGym2021}. The simulation is performed at a frequency of 120Hz, while the policies operate at a frequency of 30Hz. 4096 environments are simulated in parallel on a single A100 GPU.
A 128D latent space is used for the skill embedding. The policy, value function, and discriminator are modeled using separate multi-layer perceptrons with ReLU units and hidden layers containing $[1024, 1024, 512]$ units.
Each policy is trained using proximal policy optimization with about 7 billion samples \citep{PPO2017}, corresponding to approximately 7 years of simulated time, which requires about 2.5 days of real-world time. Selecting a weight $\lambda^{\text{task}}$ for the task reward that effectively balances the task and skill reward can be challenging, and may require task-specific tuning. We therefore apply an adaptive method to dynamically adjust $\lambda^{\text{task}}$ based on a target task-reward value \citep{compressionProportionalController}. More details are available in Appendix~\ref{app:adaptiveTask}.

\vspace{-0.1cm}
\section{Results}
We first train policies without auxiliary tasks to evaluate the model's ability to reproduce skills from a motion dataset. Examples of the policy's behaviors when given various skill commands are available in Fig.~\ref{fig:filmstripSkills}. The policy is able to follow a variety of language commands, ranging from locomotion skills, such as walking and running, to more athletic behaviors, such as sword swings and shield bashes. Since the language encoder is built on a large CLIP model \citep{ClipRadford2021}, it exhibits some robustness to new commands, which were not in the dataset. For example, the model correctly performs a casual walking motion when prompted with: \emph{``take a leisurely stroll''}, even though no captions in the dataset contained \emph{``leisurely''} or phrased walking as \emph{``taking a walk''}. However, due to the relatively small amount of captions used to train the encoder, the model can still produce incorrect behaviors for some new commands. The character successfully performs a right slash when given the prompt: \emph{``right slash''}. However, \emph{``right slash with sword''} leads the character to perform a left slash.

In addition to learning skills from a motion dataset, our policies can also be trained to perform additional high-level tasks, as outlined in Section~\ref{sec:tasks}. Examples of the tasks are available in Figure~\ref{fig:filmstripSkills}. Separate policies are trained for each task, which can then be integrated into a single multi-task controller that activates the appropriate policy given a task command. We demonstrate the effectiveness of the multi-task controller in an environment containing multiple objects that the character can interact with. The user can issue a task command for specifying the target object and the desired task that the character should perform. Our multiple-choice question-answering framework is able to consistently identify the correct task and target object from a user's commands. For example, given the command: \emph{``knock over the blue block''}. the selection model correctly identifies the policy for the Strike task, and selects the blue block as the target. 
The selection model can also parse more unusual commands, such as "mosey on down to the maroon saloon", which correctly identifies the Location task and selects the red block. Despite the generalization capabilities of large language models, some commands can still lead to incorrect behaviors. More examples of task commands and the resulting behaviors from the model are available in Appendix~\ref{sec:multiple-choice-outputs}.

\subsection{Dataset Coverage}

To determine the impact of learning a skill embedding that aligns motions and text, we evaluate our model's ability to reproduce various motions in the dataset when given the respective commands. We conduct this evaluation using a thresholded coverage metric. Given a sequence of states specified by a motion clip $\hat{\rvm} = (\hat{\rvs}_0, \hat{\rvs}_2, ..., \hat{\rvs}_n)$, a policy trajectory $\tau = (\rvs_0, \rvs_2, ..., \rvs_k)$ for a skill encoding $\rvz = \mathrm{Enc}_l(c)$ (where $c$ is a caption for $\hat{\rvm}$), and a threshold parameter $\eps > 0$, we define the coverage to be:
\begin{align}
    \text{coverage}(\tau, \hat{\rvm}, c, \eps) = \frac{1}{n} \sum_{i = 0}^{n} \mcI \jjpar{ \jjpar{ \min_{j \in \set{0, ..., k}} \tnorm{\hat{\rvs}_i - \rvs_j} } \le \eps}
    \label{eqn:coverage}
\end{align}
This metric determines the fraction of the states in a motion clip that are sufficiently close to a state in the policy's trajectory.
In our experiments we collect 300 timesteps (10 seconds) per trajectory. Instead of selecting a fixed threshold $\eps$, we apply Equation~\ref{eqn:coverage} with different values of $\eps$ between $[0, 3]$ to produce a coverage curve. 

\begin{figure}
     \centering
    \includegraphics[width=0.9\columnwidth]{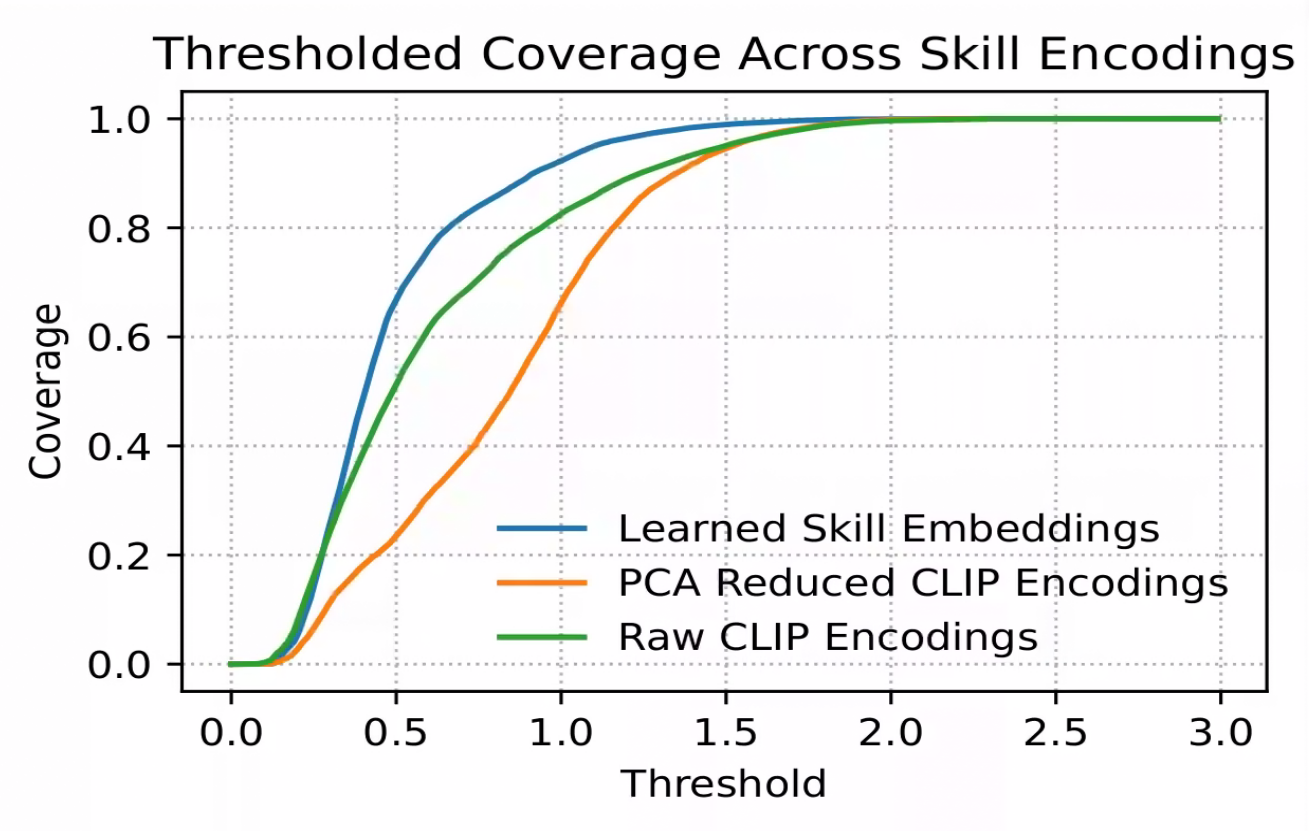}
    \vspace{-0.4cm}
    \caption{Comparing dataset coverage when different skill encodings are used during the Policy Training stage. ``Learned Skill Embeddings'' use the 128D embedding from the learned motion encoder detailed in Section~\ref{sec:skill-embedding}. We compare against baselines where policies are trained directly using the 512D CLIP text encodings of the dataset captions and where these encodings are reduced to 128D using PCA.} 
    \vspace{-0.5cm}
    \label{fig:coverage}
\end{figure}

Figure~\ref{fig:coverage} compares the performance of the PADL model with baseline models that directly use the CLIP encoding of a caption as input to the policy. Coverage statistics are averaged across all the captions for each motion clip in the dataset, and then averaged across all motion clips. The raw CLIP encoding is 512D, while our learned skill embedding is 128D. We include an additional baseline model, which uses PCA to reduce the dimensionality of the CLIP encoding to 128D. Our learned embedding is able to better reproduce behaviors in the dataset. Directly using the CLIP encoding as input to the policy tends to result in lower quality motions, and has a higher tendency of performing incorrect behaviors when directed with language commands.

\subsection{Skill Interpolation}

\begin{figure}
     \centering
     \begin{subfigure}
     \centering
         \includegraphics[width=0.8\columnwidth]{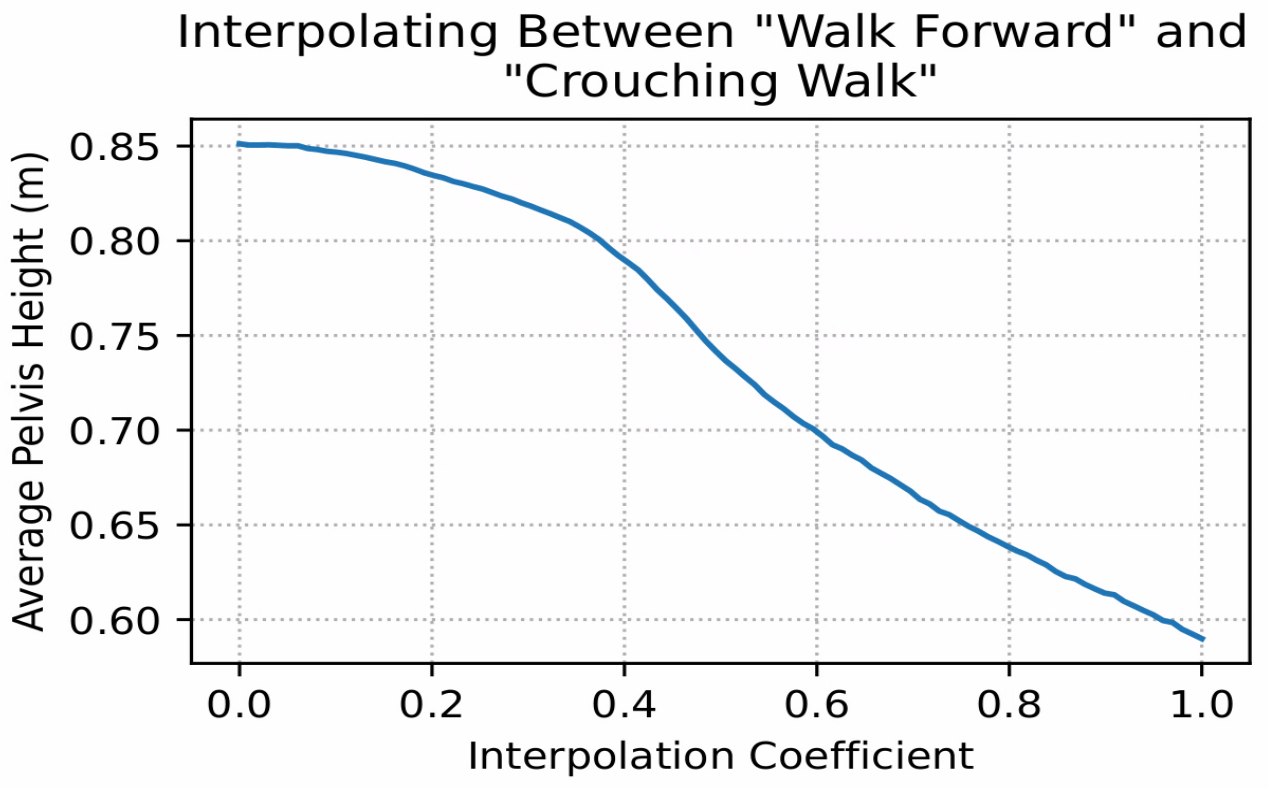}
     \end{subfigure}
     \\
     \begin{subfigure}
     \centering
         \includegraphics[width=0.8\columnwidth]{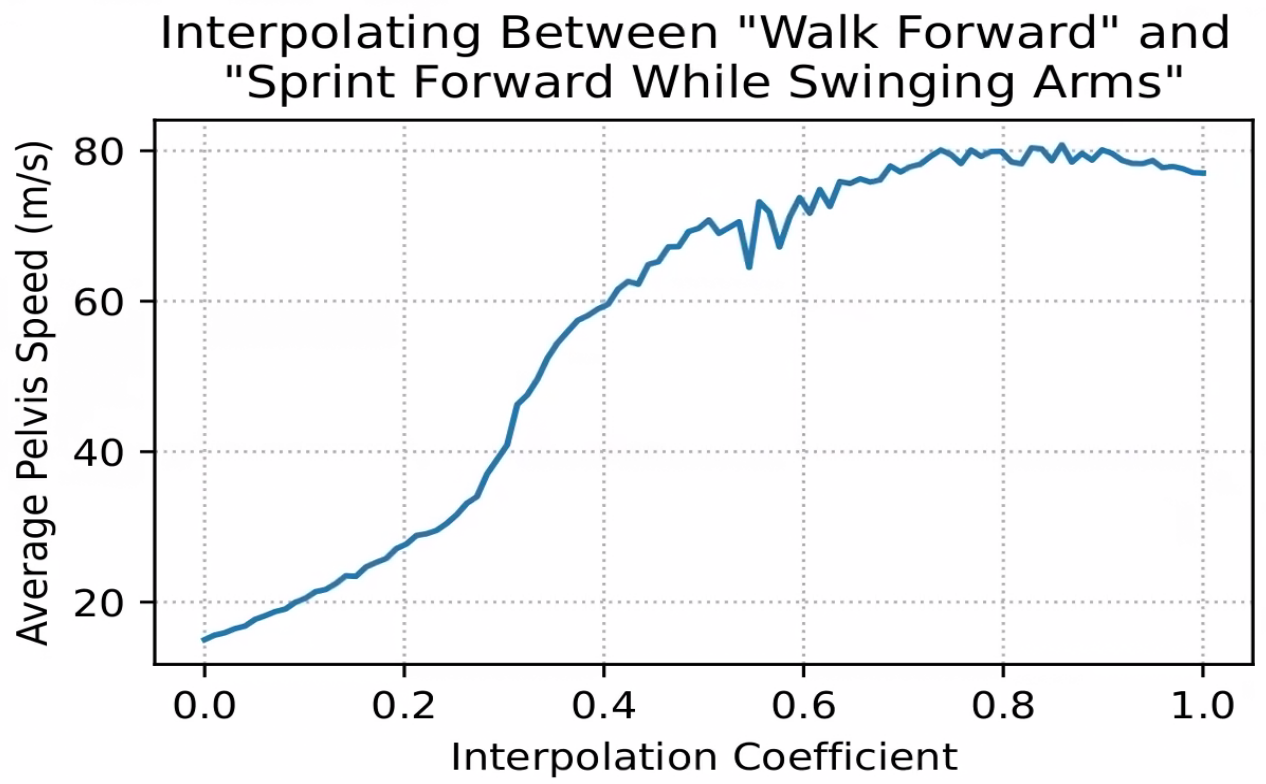}
     \end{subfigure}\\
    \vspace{-0.3cm}
        \caption{Interpolating skills in the latent space leads to semantically meaningful intermediate behaviors, such as traveling with different walking heights and speeds.}
\label{fig:slerp}
\vspace{-0.5cm}
\end{figure}

In addition to enabling language control, the learned skill embedding also leads to semantically meaningful interpolations between different skills. Given two skill commands $c_1$ and $c_2$, we encode each caption into the corresponding latents $\rvz_1$ and $\rvz_2$ using the language encoder. We then interpolate between the two latents using spherical interpolation, and condition the policy on the interpolated latent to produce a trajectory. For example, given two commands: \emph{``walk forward''} and \emph{``sprint forward while swinging arms''}, interpolating between the two latents leads to locomotion behaviors that travel at different speeds. Figure~\ref{fig:slerp} records the average velocity of the character when the policy is conditioned on different interpolated latents. Similarly, interpolating between \emph{``walk forward''} and \emph{``crouching walk forward''} leads to gaits with different walking heights. However, not all pairs of commands lead to intuitive intermediate behaviors.

\section{Conclusions}

In this work we presented PADL, a framework for learning language-directed controllers for physics-based character animation. Language is used to specify both high-level tasks that a character should perform and low-level skills that the character should use to accomplish the tasks.
While our models are able to imitate a diverse array of skills from motion data, the models remain limited in the variety of high-level tasks that they can perform.
We are interested in exploring more scalable approaches to modelling character interactions with the environment, replacing the finite \textit{a priori} collection of tasks with a more general strategy that allows the user to specify arbitrary environment interactions with natural language. We are additionally interested in scaling PADL to much larger labelled motion capture datasets \citep{BABEL:CVPR:2021}, which may lead to agents and language encoders that can model a greater diversity of skills while being more robust to paraphrasing and capable of generalizing to new commands. In particular, we expect the language encoder from the Skill Embedding stage to improve significantly with more text data. We are excited for further advances in language-guided physics-based character animation and hope that our work contributes towards the development of powerful, high-quality animation tools with broadly accessible, versatile, and easy-to-use interfaces.

\begin{acks}

We would like to thank Reallusion\footnote{\url{https://actorcore.reallusion.com/}} for providing motion capture reference data for this project. Additionally, we would like to thank the anonymous reviews for their feedback, and Steve Masseroni and Margaret Albrecht for their help in producing the supplementary video.

\end{acks}

\bibliographystyle{ACM-Reference-Format}
\bibliography{main}

\clearpage 
\pagebreak

\appendix

\section{State and Action Representation}

We evaluate the effectiveness of our framework by training language-directed control policies for a 3D simulated humanoid character. The character is equipped with a sword and shield, similar to the one used by \citet{2022-TOG-ASE}, with a total of 37 degrees-of-freedom. The character's state $\rvs_t$ is represented by a collection of features that describes the configuration of the character's body. The features include:
\begin{itemize}
    \item Height of the root from the ground.
    \item Rotation of the root in the character's local coordinate frame.
    \item Local rotation of each joint.
    \item Local velocity of each joint.
    \item Positions of the hands, feet, sword and shield in the character's local coordinate frame.
\end{itemize}

The root is designated to be the pelvis. The character's local coordinate frame is defined with the origin located at the character's pelvis, and the x-axis aligned along the root link's facing direction, with the y-axis aligned with the global up vector. The rotation of each joint is encoded using two 3D vectors, which represent the tangent and normal of the link's local coordinate frame expressed in the link's parent coordinate frame \citep{2021-TOG-AMP}. Each action $\rva_t$ specifies target rotations for PD controllers positioned at each joint. Following \citet{2021-TOG-AMP}, the target rotations for 3D joints are specified using a 3D exponential map \citet{ExpMapGrassia1998}.

\section{Task Details}
\label{app:tasks}

\subsection{Facing Task}

The facing task reward is given by:
\begin{align}
    r^{\text{task}}_t = \min\jjpar{\rvd_t \cdot \rvd^*_t, 0.5}
\end{align}
where $\rvd_t$ is the agent's facing direction. We threshold the reward, which creates an optimal ``cone" where the task reward is saturated, allowing the agent to deviate slightly from the target heading in order to better imitate skills.

\subsection{Location Task}
The location task reward is calculated according to:
\begin{align}
    r^{\text{task}}_t = \jjcases{ 0.2 r_t^{\text{pos}} + 0.8r^{\text{vel}}_t & \tnorm{{\rvx}^* - {\rvx}} > \delta_{\text{pos}} \\ 0.8  & \tnorm{{\rvx}^* - {\rvx}} \le \delta_{\text{pos}}  }
\end{align}
where $\rvx$ denotes the position of the character's root, and $r^\text{pos}_t$ encourages the character to be close to the target:

\begin{align}
    r_t^{\text{pos}} &= \exp\jjpar{-0.25\tnorm{\rvx^* - \rvx}^2},
\end{align}
$r^\text{vel}_t$ encourages the character to move towards the target. This velocity reward incentivizes the agent to travel speed of at least $\delta_\text{vel} = 0.5$ m/s in the direction of the target, and not travel in any other direction:
\begin{align}
    r^{\text{vel}}_t &= \exp\jjpar{-0.25\jjpar{\max(\delta_\text{vel} - v_t^\text{proj}, 0) + 0.1v_t^\text{perp}}}
\end{align}
where 
\begin{align}
    v_t^\text{proj} &= \tnorm{\text{proj}_{\rvx^*}(\rvv_t)} \\
    v_t^\text{perp} &=  \tnorm{\text{perp}_{\rvx^*}(\rvv_t)}
\end{align}
define the agent's velocity in the direction of and tangent to the target, respectively. We saturate the task reward when the agent gets within $\delta_{\text{pos}} = 2$m of the target, and terminate the episode when the block is knocked over to disincentivize the agent simply running into the block.

\subsection{Strike Task}

Finally, we have a strike task, where the objective is for the character to knock over a target object. The goal $\rvg_t = (\tilde{\rvx}^*_t, \tilde{\dot{\rvx}}^*_t, \tilde{q}^*_t, \tilde{\dot{q}}^*_t)$ records the target object's position $\tilde{\rvx}^*_t$, rotation $\tilde{q}^*_t$, linear velocity $\tilde{\dot{\rvx}}^*_t$, and angular velocity $\tilde{\dot{q}}^*_t$. All features are expressed in the character's local coordinate frame. The task-reward is then given by:
\begin{align}
    r^{\text{task}}_t = \jjcases{  0.2 r_t^{\text{pos}} + 0.8 r^{\text{vel}}_t + 0.8 r_t^{\text{knock}}  & u_t^* \cdot u^{\text{up}} \ge 0.3 \\ 1.4 &  u_t^* \cdot u^{\text{up}} < 0.3 }
\end{align}
where the knock reward incentivizes the agent to knock over the block:
\begin{align}
    r_t^{\text{knock}} &= 1 - u_t^* \cdot u^{\text{up}}.
\end{align}
Here, $u^{\text{up}}$ is the global up vector, and $u_t^*$ is target object's local up vector expressed in the global coordinate frame. The position reward $r_t^{\text{pos}}$ and velocity reward $r^{\text{vel}}_t$ are the same as those used for the location task. The task reward saturates when the block has been sufficiently tipped over. \\

\subsection{Adaptive Task Weight Schedule}
\label{app:adaptiveTask}
Selecting a weight $\lambda^{\text{task}}$ for the task reward that effectively balances the task and skill reward can be challenging, and can require task-specific tuning.
Setting $\lambda^{\text{task}}$ too low can lead to policies that only learn to imitate skills without any regard for the task. Similarly, when $\lambda^{\text{task}}$ is too high, the policy can learn to perform the task using unnatural behaviors, entirely ignoring the skill command. Therefore, instead of using a constant task weight or manually constructing an annealing schedule for $\lambda^{\text{task}}$, we use a proportional controller to dynamically adjust $\lambda^{\text{task}}$ over the course of the training process, in a similar manner as \citet{compressionProportionalController}. The controller is parameterized by a target task reward $\hat{r}^\text{tar}$, as well as by a controller gain $k_p$ and a small positive constant $\eps$ for numerical stability. At epoch $i$, we calculate the mean task reward $\ov{r}^{\text{task}}_i$ across the experience buffer. We then update the task weight $\lambda^\text{task}_i$ according to the error between $\ov{r}^{\text{task}}_i$ and $\hat{r}^\text{tar}$ in log-space:
\begin{align}
    \lambda^\text{task}_{i+1} &= \exp\jjpar{ \log\left(\lambda^\text{task}_i\right) + k_p\jjpar{\log \left(\hat{r}^\text{tar} + \eps \right) - \log \left(\ov{r}^{\text{task}}_i + \eps \right)} }
\end{align}
The task weight is initialized to be $\lambda^\text{task}_0 = 3$, and $\lambda^\text{task}_i$ is clamped to the range $[0.5, 3]$. For the location task we set a target task reward weight of 0.15, while for the strike task we set a target reward of 0.3. For the facing task we found the controller to be unnecessary and used a constant $\lambda^\text{task}= 1$.

\begin{figure*}[t]
	\centering
    \subfigure["forward walk", "walk forward while swaying arms".]{\includegraphics[width=\linewidth]{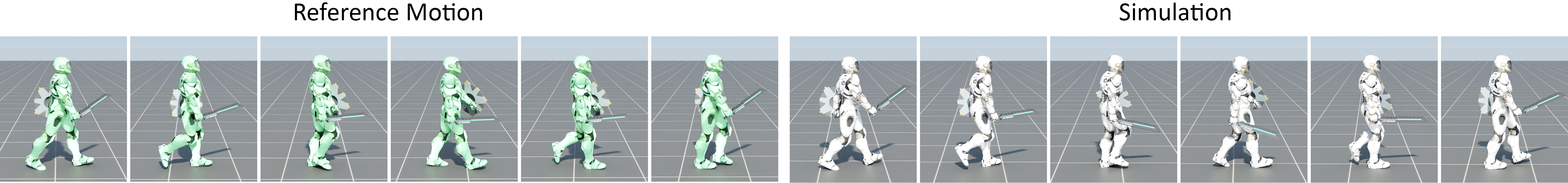}}\\
    \subfigure["sprint forwards while swinging arms".]{\includegraphics[width=\linewidth]{figures/filmstrip_run.png}}\\
    \subfigure["kick", "kick with right leg", "right kick into step forward", "right leg kick".]{\includegraphics[width=\linewidth]{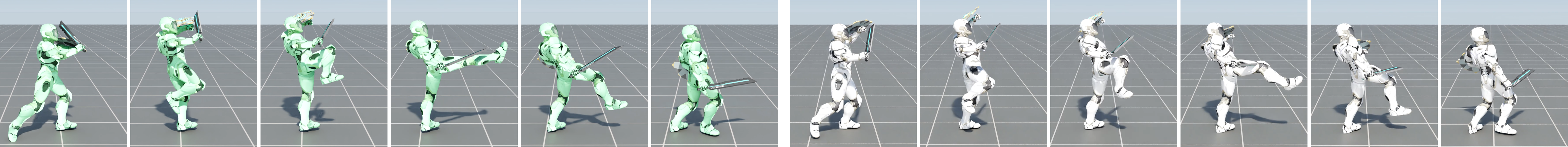}}\\
    \subfigure["left shield bash", "shield bash left", "shield bash to the left while standing still".]{\includegraphics[width=\linewidth]{figures/filmstrip_shield_bash.png}}\\
    \subfigure["slash right", "right swing", "swing sword to the right", "stand still and slash to the right".]{\includegraphics[width=1\linewidth]{figures/filmstrip_sword_swing.png}}\\
    \caption{Reference motion clips (left side) and their corresponding captions, along with motions produced by a simulated character when directed to perform the reference skills through language commands (right side).}
    \label{fig:moreFilmstripSkills}
\end{figure*}

\section{Multiple-Choice Model Example Outputs}
\label{sec:multiple-choice-outputs}

In Table \ref{tab:commands}, we provide examples of task commands and the corresponding object and policy that the multiple-choice QA model identified. We observe that the QA model is able to correctly identify the user's intent even when provided with exotic task commands such as ``destroy the green guy'' or ``mosey on down to the maroon saloon''. We also provide several examples where the QA model incorrectly identifies the task and/or object. For example, the model predicts that the task command ``go to the blue target'' references the strike task instead of the location task, while ``go to the blue block'' and ``go to the blue tower'' are correctly identified as the location task. The QA model is also occasionally sensitive to paraphrasing, such as when it correctly identifies the task in ``navigate to the lime rectangular prism" but not in ``navigate toward the lime rectangular prism".

\begin{table*}[]
    \centering
    \caption{Example task commands and the corresponding object and task identified by the multiple-choice QA model.}
    \begin{tabular}{|c|c|c|}
    \hline \textbf{Task Command} & \textbf{Identified Object} & \textbf{Identified Task} \\
\hline "knock over the blue block" & "the blue object nearby." \cmark & "knock over a specified object." \cmark\\
\hline "knock over the green block" & "the green object nearby." \cmark & "knock over a specified object." \cmark\\
\hline "go to the red block" & "the red object nearby." \cmark & "navigate to a specified destination."\cmark \\
\hline "go to the orange block" & "the orange object nearby."\cmark & "navigate to a specified destination."\cmark \\
\hline "face the purple block" & "the purple object nearby." \cmark& "orient himself to face a specified heading."\cmark \\
\hline "knock over the purple target" & "the purple object nearby." \cmark& "knock over a specified object." \cmark\\
\hline "turn towards the blue target" & "the blue object nearby." \cmark & "orient himself to face a specified heading."\cmark \\
\hline "turn towards the orange target" & "the orange object nearby." \cmark& "orient himself to face a specified heading."\cmark \\
\hline "face the orange target" & "the orange object nearby." \cmark& "orient himself to face a specified heading."\cmark \\
\hline "face the purple target" & "the purple object nearby." \cmark& "orient himself to face a specified heading."\cmark \\
\hline "go to the blue target" & "the blue object nearby." \cmark & "knock over a specified object." \xmark \\
\hline "topple the red tower" & "the red object nearby."\cmark & "knock over a specified object." \cmark\\
\hline "face the orange obelisk" & "the orange object nearby." \cmark & "orient himself to face a specified heading." \cmark \\
\hline "navigate to the lime rectangular prism" & "the green object nearby." \cmark & "navigate to a specified destination." \cmark \\
\hline "navigate toward the lime rectangular prism" & "the green object nearby." \cmark & "orient himself to face a specified heading." \xmark \\
\hline "look at the stop sign" & "the red object nearby." \cmark & "orient himself to face a specified heading." \cmark \\
\hline "watch the sunset" & "the red object nearby." \cmark& "orient himself to face a specified heading." \cmark\\
\hline "knock over the cobalt block" & "the red object nearby." \xmark & "knock over a specified object." \cmark \\
\hline "get close to the violet marker" & "the purple object nearby."\cmark & "orient himself to face a specified heading." \xmark \\
\hline "destroy the green guy" & "the green object nearby." \cmark & "knock over a specified object." \cmark \\
\hline "mosey on down to the maroon saloon" & "the red object nearby."\cmark & "navigate to a specified destination." \cmark\\

    \hline
    \end{tabular}
    \label{tab:commands}
\end{table*}

\section{Comparing PADL to Other Adversarial RL Frameworks}
\label{sec:mode-collapse}

When training PADL agents (detailed in Section \ref{sec:policy-training}), the skill objective explicitly rewards agents for being able to imitate every motion clip in the dataset, using a discriminator trained on the joint distribution of state transitions and skill embeddings. We find that the use of a joint discriminator helps to mitigate mode collapse during PADL training when compared to other work in adversarial reinforcement learning that uses discriminators trained only on the marginal distribution of state transitions. Here we specifically compare our method to two related adversarial RL frameworks, AMP \citep{2021-TOG-AMP} and ASE \citep{2022-TOG-ASE}.

\subsection{Comparison to AMP}

AMP, like PADL, trains agents using a combination of task and skill rewards. However, since AMP's skill reward uses a marginal discriminator, mode collapse can occur, where agents focus on imitating a specific subset of skills in the reference motion data while completing the high-level task. PADL's use of a joint discriminator in the skill reward, where policies are explicitly trained to accomplish the high-level task using different reference skills, can improves a policy’s coverage of the dataset. Moreover, PADL agents, unlike AMP agents, are conditioned on a latent variable encoding the skill to be used. This allows a user to control in real-time which skills a trained agent uses to accomplish a task, which is crucial for effective language control.

\subsection{Comparison to ASE}

Both ASE low-level controllers and PADL controllers are conditioned on skill latents, allowing the skill the agent uses to be dynamically controlled. During ASE training, latents are drawn randomly from a prior distribution (e.g. the unit sphere); the policy learns a meaningful representation on this latent space throughout training using a marginal discriminator combined with an encoder that promotes high mutual information between a latent and its corresponding policy trajectory. This approach too can lead to mode collapse, with only a subset of skills from the reference dataset being represented in the latent space. PADL mitigates this type of mode collapse by assigning a distinct motion latent to every motion clip in the reference dataset (these latents are learned in the Skill Embedding stage), guaranteeing that every motion clip is represented in the latent space.

In one of our early experiments developing language-controlled animation systems, we attached a language head on top of an ASE low-level controller. We created a dataset of (latent, caption) pairs by sampling latents from the unit sphere, recording trajectories from a pre-trained controller checkpoint with those latents, and annotating the trajectories with natural language. We then trained a small MLP to reverse the annotation process and map the BERT embeddings of a trajectory's caption to the corresponding latent that produced the trajectory. This approach allowed for a policy's skill to be controlled with language, but is annotation inefficient, since each dataset of (latent, caption) pairs is only applicable for a specific checkpoint’s learned latent space. A different ASE checkpoint (which possesses a different learned latent space) requires the collection of an entirely new dataset of annotations. Moreover, due to mode-collapse, more complicated skills in the dataset were often not represented in the policy's latent space.

\begin{algorithm}[t]
\SetAlgoNoLine
agentState $\leftarrow$ agent state\;
\While{not done}{
    skillLatent = $\mathrm{Enc_l}$(getSkillCommand())\;
    policyIdx, objectIdx = QAModel(getTaskCommand())\;
    policy = policies[policyIdx]\;
    targetObjectState = objects[objectIdx]\;
    action = policy(agentState, skillLatent, targetObjectState)\;
    agentState = env.step(action)\;
}
\caption{Multi-Task Aggregation}
\label{alg:multi-task}
\end{algorithm}

\end{document}